\documentclass{article}
\usepackage{graphicx}
\usepackage{algorithm2e}

\usepackage{xcolor}
\def\qq#1{{\color{black}{#1}}}

\def\be{\begin{equation}}
\def\ee{\end{equation}}
\newcommand{\ff}[1]{{\bf  #1}}
\def\a{\alpha}
\def\b{\beta}
\def\lam{\lambda}
\def\x{\ff{x}}

\begin{document}

\title{Multi-Species Cuckoo Search Algorithm for Global Optimization}

\author{Xin-She Yang$^1$, Suash Deb$^2$, Sudhanshu K Mishra$^3$ \\[15pt]
1) School of Science and Technology, 
Department of Design Engineering and Mathematics,\\
Middlesex University,  London NW4 4BT, UK. \\[10pt]
2) IT \& Educational Consultant, Ranchi, India, and 
Distinguished Professorial Associate, \\
Decision Sciences and Modelling Program, 
Victoria University, Melbourne, Australia. \\[10pt]
3) Department of Economics, 
North-Eastern Hill University, Shillong, India.
}

\date{ }

\maketitle

\noindent {\bf Citation Details:} Xin-She Yang, Suash Deb, Sudhanshu K Mishra, Multi-species cuckoo search algorithm for global optimization, Cognitive Computation, vol. 10, number 6,
1085-1095 (2018). \\[15pt]

\centerline{Abstract}
\hrule
\begin{itemize}

\item[] Background: \qq{Many optimization problems in science and engineering are highly nonlinear, and thus require sophisticated optimization techniques to solve. Traditional techniques such as gradient-based algorithms are mostly local search methods, and often struggle to cope with such challenging optimization problems. Recent trends tend to use nature-inspired optimization algorithms. }

\item[] Methods: \qq{This work extends the standard cuckoo search (CS) by using the successful features of the cuckoo-host co-evolution with multiple interacting species, and the proposed multi-species cuckoo search (MSCS) intends to mimic the multiple species of cuckoos that compete for the survival of the fittest, and they co-evolve with host species with solution vectors being encoded as position vectors. The proposed algorithm is then validated by 15 benchmark functions as well as five nonlinear, multimodal design case studies in practical applications.}

\item[]  Results: \qq{Simulation results suggest that the proposed algorithm can be effective for finding optimal solutions and in this case all optimal solutions are achievable. The results for the test benchmarks are also compared with those obtained by other methods such as the standard cuckoo search and genetic algorithm, which demonstrated the efficiency of the present algorithm.}

\item[] Conclusions: Based on numerical experiments and case studies, we can conclude that the proposed algorithm can be more efficient in most cases, leading a potentially very effective tool for solving nonlinear optimization problems.
\end{itemize}


\hrule 

\section{Introduction}

Many applications involve optimization, and thus require sophisticated optimization algorithms to solve. Such applications can be very diverse, spanning many areas and disciplines from engineering designs and scheduling to data mining and machine learning \cite{YangDeb,YangDebRev,Dubey,Sidd,Wu}. One of the current trends is to use metaheuristic algorithms inspired by the successful characteristics in nature. Among these new algorithms, cuckoo search has been shown to be powerful in solving many problems \cite{YangDeb,YangDebRev}. This standard version was mainly designed for single objective optimization problems, which was later extended to multiobjective optimization \cite{YangDeb2013}. Both versions used simplified characteristics to represent the brood parasitism of some cuckoo species and their interactions with host birds species.

However, the reality is far more complicated in the cuckoo-host co-evolution systems \cite{Davi}. \qq{The co-evolution often involves multiple cuckoo species that compete with each other and compete for the resources of host bird species, while the hosts can also have multiple species. Cuckoo species tend to evolve to lay eggs with mimicry of the size, colours and texture of the eggs of host birds, often with critical timing advantage, while host birds
can counter-act such parasitism by developing their defensive strategies to identify potential intruding eggs and minimize the risk of hatching cuckoo eggs. This arms race is both co-evolutionary and ongoing \cite{Davi2,Payne}, and the co-evolution seems to promote species richness and subspecies variations as well as diversity in parasitic cuckoos \cite{Kruger}. }

One way to model such co-evolution has been proposed by Mishra \cite{Mishra} to use a host-parasite co-evolutionary approach where both parasites and hosts took random flights and the probability of detection or rejection of eggs is dynamic, depending on the accumulative success of the parasites such as cuckoos. This approach has been used to solve both function optimization problems and completing incomplete correlation matrix \cite{Mishra}. However, this approach only captured a very small part of the major characteristics of the cuckoo-host co-evolutionary systems.

\qq{In order to capture more detailed characteristics of this co-evolution system, in this paper, we extend the original cuckoo search to a new multi-species co-evolutionary cuckoo search algorithm, which simulates the main co-evolutionary behaviour of both cuckoo species and host species.
Therefore, the paper is organized as follows. Section 2 summarizes the original cuckoo search and its main equations. Section 3 outlines the novel features of the new multi-species cuckoo search, followed by the numerical experiments on 15 different test benchmarks in Section 4. Section 5 presents the results of five different case studies concerning engineering designs, inverse parameter estimation and data clustering. The paper concludes with discussions about further research directions in Section 6. }


\section{The Original Cuckoo Search}

Cuckoo search (CS) is a nature-inspired metaheuristic algorithms,
developed in 2009 by Xin-She Yang and Suash Deb \cite{YangDeb}.
CS is based on the brood parasitism of some cuckoo species. In addition, this
algorithm is enhanced by the so-called L\'evy flights \cite{Pav},
rather than by simple isotropic random walks.
Recent studies show that CS is potentially far more efficient than PSO
and genetic algorithms \cite{YangDeb2010,Gandomi}. A relatively comprehensive review of the
studies up to 2014 was carried out by Yang and Deb \cite{YangDebRev}.

\subsection{Cuckoo Search and its Algorithmic Equations}

In the natural world, many cuckoo species (59 species among 141 cuckoo species) engage the so-called obligate reproduction parasitism strategy. There is an evolutionary arms race between such cuckoo species and their
associated host species \cite{Davi,Davi2,Payne}. Based on such phenomena, Yang and Deb developed the
cuckoo search in 2009 \cite{YangDeb}, which uses three simplified rules: 1) Each cuckoo lays one egg at a time, and dumps it in a randomly chosen nest. 2) The best nests with high-quality eggs will be carried over to the next generations. 3) The number of available host nests
is fixed, and the egg laid by a cuckoo is discovered by the host bird
with a probability $p_a \in [0,1]$. In this case, the host bird can either
get rid of the egg, or simply abandon the nest and build a completely new nest at a new location.

In the original cuckoo search,  there is no distinction between an egg, a nest, or a cuckoo,
as each nest corresponds to one egg which also represents one cuckoo, which makes it much easier to implement \cite{YangDeb2010}. Mathematically speaking, cuckoo search uses a combination of a local random walk and the global explorative random walk,
controlled by
a switching parameter $p_a$. The local random walk can be written as
\be \x_i^{t+1}=\x_i^t +\beta s \otimes H(p_a-\epsilon) \otimes (\x_j^t-\x_k^t), \label{equ-100}  \ee
where $\x_j^t$ and $\x_k^t$ are two different solutions selected randomly by random permutation,
$H(u)$ is a Heaviside function, $\epsilon$ is a random number drawn from a uniform distribution, and
$s$ is the step size. Here $\beta$ is the small scaling factor.

On the other hand, the global random walk is carried out by using L\'evy flights
\be \x_i^{t+1}=\x_i^t+\a \otimes L(s, \lam), \label{equ-200} \ee
where \be L(s, \lam) \sim \frac{\lam \Gamma(\lam) \sin (\pi \lam/2)}{\pi}
\frac{1}{s^{1+\lam}}, \quad (s \gg 0), \label{Levy-eq-100} \ee
 \qq{where $\a>0$ is the step size scaling factor, which should be related to the scales of the problem of interest. Here `$\sim$' highlights the fact that the search steps in terms of  random numbers $L(s,\lam)$
should be drawn from the L\'evy distribution on the right-hand side of Eq.~(\ref{Levy-eq-100}), which approximates the L\'evy distribution by a power-law distribution with an exponent $\lam$. In addition, here $\otimes$ denotes that the multiplication is an entry-wise operation.  The use of L\'evy flights makes the algorithm more likely to jump out of any local optima, enabling a better exploration ability  \cite{Pav,YangDeb,YangDebRev}. }

\subsection{Cuckoo Search in Applications}

Since the development of the cuckoo search algorithm in 2009, it has been applied
in many areas, including optimization, engineering design, data mining and computational intelligence with promising efficiency. From the case studies in engineering
design applications, it has been shown that cuckoo search has superior performance to other algorithms for a range of continuous optimization problems \cite{YangDeb2010,Gandomi,Fister,YangDebRev,Yildiz}. A review by Yang and Deb covered the literature up to 2013 \cite{YangDebRev}, while a review by Fister et al. covered the literature up to 2015, and another review by Mohamad et al. focused on applications up to 2014 \cite{Moham}. The most recent literature has been carried out by Shehab et al. \cite{Shehab}, which covers some of the literature up to 2017. These reviews have briefly outlined some of the diverse applications using cuckoo search and its variants.

Other applications include vehicle component optimization \cite{Durgun}, wireless sensor networks \cite{Dhiv}, training neural networks \cite{Valian}, runoff-erosion modelling \cite{Santos},  phase equilibrium in thermodynamic calculations \cite{Bhar},
network optimization \cite{Mora}, scheduling  \cite{Chand},
multilevel color image segmentation \cite{Pare},
biodiesel engine optimization \cite{Wong},
graphic objective feature extraction \cite{Woz}, fractional order PID control design \cite{Zamani}, vulnerabilities mitigation \cite{Zine} and others \cite{YangCSFA}.

In addition, different cuckoo search variants have been developed. For example, a binary cuckoo search for
feature selection has been developed by Pereira et al. \cite{Pereira}.
Walton et al. \cite{Walton} developed a modified cuckoo search for solving complex mesh generation in engineering simulation, while Zheng and Zhou \cite{Zheng} provided
a variant of cuckoo search using Gaussian process.
Mlakar et al. developed a self-adaptive cuckoo search \cite{Mlak}, while Wang et al.
enhanced cuckoo search with chaotic maps \cite{Wang}.

As a further extension, Yang and Deb \cite{YangDeb2013} developed a multiobjective cuckoo search (MOCS) algorithm for engineering design applications.
Recent studies have demonstrated that cuckoo search can perform significantly better than other algorithms in many applications \cite{Gandomi,YangCSFA,Yildiz}.

\section{Multi-Species Cuckoo Search}
In the original cuckoo search and its many variants, there is only one cuckoo species
interacting with one species of host birds. \qq{ In the standard cuckoo search, a cuckoo is allowed to lay a single egg and each nest contains only a single egg. This is a very simplified scenario. In the real-world cuckoo-host systems, it is observed that multiple cuckoo species co-evolve with multiple host species to compete for survival of the fittest by brood parasitism \cite{Davi,Payne}. Loosely speaking, cuckoos can evolve to subspecies with speciation, and they can be subdivided into different gentes
targeting at different host species \cite{Davi2,Kruger}. The interaction dynamics can be very complex, forming an on-going, co-evolutionary arms race between cuckoo subspecies and host species as well as different cuckoo species. Studies have shown that such co-evolution may promote species richness in parasitic cuckoos with the enhanced speciation and extinction rates \cite{Kruger}. Strictly speaking, different species, subspecies and gentes are
used in the biological literature and their meanings are different \cite{Kruger,Payne}; however, we will simply use species here for simplicity. }

Based on these characteristics, we can extend the original cuckoo search to capture more realistic cuckoo-host co-evolution to develop a new multi-species co-evolution cuckoo search, or multi-species cuckoo search (MSCS) for short.

In order to describe the multi-species cuckoo search in detail, we use the following idealized rules/assumptions:
\begin{enumerate}
\item There are multiple cuckoo species ($m$ species) that compete and co-evolve with
a host species. The arms race between cuckoo species and the host species obeys
the survival of the fittest. Both the best cuckoos and the best hosts will pass onto the next generation.

\item Each cuckoo of any cuckoo species can lay $r$ eggs inside a randomly selected
host nest. Each cuckoo egg has a probability $p_a$ to be discovered (and then abandoned)
by the host (thus the survival probability is $1-p_a$).

\item Each nest contains $q$ eggs. If the fraction/ratio of cuckoo eggs is higher than $1-p_a$, the host bird can abandon the nest and then fly away to build a new nest in a new location.
\end{enumerate}

Based on these rules for MSCS, we can represent them more mathematically. Each solution $\x_i$
to an optimization problem is represented by an egg, which corresponds to a
$D$-dimensional vector. Therefore, an egg is equivalent to a solution,
and a nest represents a group of $q$ solutions.

In general, there are $m \ge 1$ species with a total population of $n$, each species has
$n_i \; (i=1,...,m)$ cuckoos such that  \be \sum_{i=1}^m n_i =n. \ee
Each cuckoo lays $r \ge 1$ eggs. There are $w$ host nests, and each nest can have $q \ge 1$ eggs on average. So the total number of eggs in
the host nests are $N_h=w q$, which should be greater
than $n$. That is, $N_h \gg n$. In nature, it is estimated that
approximately $1/4$ to $1/2$ of eggs in the host nests are cuckoo eggs. Thus, we can
set $n=N_h/2$ for simplicity.

\qq{Competition can occur at three different levels: intraspecies competition, inter-species competition and cuckoo-host competition. Even for a single species,
cuckoos within the same species can compete for host nests. For multiple species,
one species of cuckoos can compete with cuckoos from other species by egg-replacing
strategy. The most significant competition is the cuckoo-host competition.
All these three kinds of competition interact and co-evolve to form
a complex system, leading to potential self-organizing intelligent behaviour.
}

\qq{
Different cuckoo species will compete for survival of the fittest, and they can take over other cuckoos' territory or replace eggs laid by other cuckoo species. This can be done simply by random swapping its location vector with another in a dimension-by-dimension manner. This binary random swapping operator can be realized by
\be \x_a^{(\textrm{new})}= \x_a \otimes (1-\ff{Q}) + \x_b \otimes \ff{Q}, \label{swap-100} \ee
and
\be \x_b^{(\textrm{new})} = \x_a \otimes \ff{Q} + \x_b \otimes (1-\ff{Q}), \label{swap-200} \ee
where $\x_a$ is randomly selected from cuckoo species $a$, while $\x_b$ is selected from
cuckoo species $b$. Here, $Q$ is a random binary vector with the same length of $\x_a$
and each of its components is either 1 or 0 [i.e., $Q_k \in \{0, 1\} \; (k=1, 2, ..., D)$]. For example, $\ff{Q}=[1, 0, 0, 1, 0, 1, 1]$ is a binary vector in a 7-dimensional space.
Again, $\otimes$ means that the operation is a component-wise or dimension-wise
operation. }

The main steps for implementing and simulating the above idealized characteristics are as follows:
\begin{enumerate}
\item \qq{There are two population sets: a total population of $n$ cuckoos for $m$ cuckoo species (each has its own population $n_j \; (j=1,2,...,m)$), and a population of $N_h$ host nests. Thus, there are two initial best solutions: $\ff{g}_{cs}^*$ to denote the best cuckoo among all $m$ cuckoo species (each species has its own best cuckoo
    $\ff{g}_j^*$) and $\ff{g}_h^*$ to denote the
    best host in terms of objective fitness. }

\item For each generation of evolution, each cuckoo (say, cuckoo $i$) from a cuckoo species (say, species $j$) can lay $r$ eggs in a randomly selected host nest (say, nest $k$). The newly laid eggs will replace randomly selected eggs in the nest (so that the total
    number of eggs in the nest remain constant $q \ge1$). The main equation for this action can be carried out by Eq.(\ref{equ-100}).

\item For any new egg laid by a cuckoo, there is a probability of $p_a$ to be discovered.
Among $q$ eggs, if the fraction of cuckoo eggs exceeds $1-p_a$, the host can abandon the nest completely and fly away to build a new nest at a new location via Eq.(\ref{equ-200}).

\item \qq{Different species of cuckoos will compete for survival and territories, thus they can lay eggs in nests that other cuckoos just laid. This competition is equivalent to replacing or swapping its own eggs with another cuckoo's eggs from different species. Thus, it can be achieved by randomly swapping their components dimension by dimension via (\ref{swap-100}) and (\ref{swap-200}). }

\item Random mixing is carried out in terms of egg-laying and nest choices among different cuckoo species and the host species.

\item Both the best cuckoos and host nests (in terms of their fitness) should pass onto the next generation.

\end{enumerate}

These key steps can be schematically represented as the pseudocode in Algorithm~\ref{alg-100}.
To illustrate the main ideas, for two species of cuckoos with a total population of $n=40$,
we have $m=2$ and $n=40$. If two species have the same population size, we have $n_1=n_2=n/2=20$. For simplicity, if all nests have the same number of four eggs in $20$ nests, we have $q=4$ and $w=20$, thus there $N_h=20*4=80$ eggs in all the nests.
In addition, if each cuckoo lays one egg at a time (or $r=1$), this means that
there are $n r =40$ cuckoo eggs in the cuckoo-host system. Thus,
in this case, there are exactly $50\%$ of the eggs belong to cuckoos in the combined population of cuckoo species and host nests.

\qq{Obviously, the number of cuckoo eggs in a particular nest can be randomly distributed
from $1$ to $q=4$. For $p_a=0.25$, if there are $3$ or $4$ eggs in a nest, one egg may typically belong to cuckoos. If the number of alien eggs is higher, this nest can be abandoned by its host, and thus a new replacement nest with $q$ new eggs (or randomly generated solutions) will be built in a new location, typically far enough from the original location. }

\begin{algorithm}

\hrule
Initialize parameters, cuckoos and host populations\;
Find the best cuckoo $\ff{g}_{cs}^*$ and best host $\ff{g}_h^*$\;
\While{stopping criterion is not met}{
Choose a cuckoo species randomly (say, species $j$)\;
Select a cuckoo in the species randomly (say, $i$)\;
Generate a random number $\epsilon$ from [0,1]\;
\eIf {$\epsilon<p_a$ (discovery probability)}
{Generate a new solution by Eq.(\ref{equ-100})\; }
{Perform a L\'evy flight by Eq.(\ref{equ-200})\;}
Put into a host nest randomly (say, nest $k$)\;
\qq{Generate a random binary vector $\ff{Q}$ of $D$ dimensions \;
Swap two randomly selected eggs from two different species by (\ref{swap-100}) and
(\ref{swap-200})\; }
Evaluate all new fitness/objectives\;
\If{New solution is better}{Replace the worst egg in nest $k$
by the new solution\;}
Update the best solution $\ff{g}_{cs}^*$ among all cuckoos\;
\If{cuckoo eggs ratio in nest $k$ is higher than $1-p_a$}
{Perform a L\'evy flight and build a new nest at a new location via Eq.(\ref{equ-200})\;}
Update the current best host $\ff{g}_h^*$ among all nests\;
\qq{Pass the best cuckoos in each species and host nests to next generation\;}
Find and record the overall best solution $\x_*$\;
Update iteration counter $t$\;
} \hrule
\caption{Multi-species co-evolutionary cuckoo search. \label{alg-100} }
\end{algorithm}

\qq{It is worth pointing out that there seems to have some similarity between multi-species cuckoo interactions and the multi-swarm optimization in the literature \cite{Blackwell}. However, there
are two key differences here: the multi-species cuckoo search (MSCS) mimics the
co-evolution between parasite cuckoo species and host species, while mutli-swarms mainly
split a population of the same kind into subgroups or subswarms. In addition, the share of information in MSCS is among the same cuckoo species and the same host species, not directly shared among competing species. Such information-sharing structure can potentially enable extensive exploitation of local information as well as global information. On the other hand, multi-swarms tend to share information among all subswarms. Furthermore, different cuckoo species compete for survival, while the multi-swarms do not compete. These differences mean that MSCS is not a simple multi-swarm system, rather it is an interacting co-evolving multi-swarm system. Therefore, different characteristics and performance can be expected.
}

\section{Validation by Numerical Experiments}

All new algorithms have to be extensively tested by a diverse range of benchmarks and case studies. As a preliminary test, we will use a subset of 15 function benchmarks and
5 case studies.

\subsection{Benchmarks}
\qq{
For this purpose, we have selected 15 benchmark functions with different modalities and objective landscapes from traditional optimization functions,
the CEC2005 test suite and most
recent CEC2015 test functions. The chosen set of benchmarks have diverse properties so that we can test the proposed algorithm more thoroughly. }

The first function is the shifted sphere function $f_1$ from the CEC2005 benchmark suite \cite{CEC2005}. This function has the global minimum $f_{1,\min}=-450$ in the domain $-100 \le x_i \le 100$.

The second function is Ackley function
\be f_2(\x)=-20 e^{-\frac{1}{5} (\frac{1}{D} \sum_{i=1}^D x_i^2)^{1/2}}
 - e^{\frac{1}{D} \sum_{i=1}^D \cos (2 \pi x_i)} +20 +e, \ee
which has its global minimum $f_*=0$ at $(0,0,...,0)$. This function is highly nonlinear and
multimodal.

The third function is Yang's forest-like function
\be f_3(\x)=\Big( \sum_{i=1}^D |x_i| \Big) \exp\Big[- \sum_{i=1}^d \sin (x_i^2) \Big],\ee
which has the global minimum $f_*=0$ at $(0,0,...,0)$ in the domain of $-2 \pi \le x_i \le 2 \pi$.
 This function is highly nonlinear and multimodal, and its first derivatives do not exist at the optimal point due to the modulus $|.|$ factor.

The fourth function is the shifted Schwefel's problem with noise in fitness as given in CEC2005 benchmark suite \cite{CEC2005}, which has the mean global minimum $f_{4,\min}=-450$ in the domain $-100 \le x_i \le 100$.

The fifth function is Schwefel's Problem 2.22 \cite{Yao}
\be f_5(\x)=\sum_{i=1}^D |x_i|+\prod_{i=1}^D |x_i|, \ee
which has the global minimum $f_{5,\min}=0$ at $\x=(0,0,...,0)$
in the domain $-10 \le x_i \le 10$. This function is unimodal.

The sixth function is the shifted Rosenbrock function $f_6$ of CEC2005 benchmark suite with the minimum $f_{6,\min}=390$ in $-100 \le x_i \le 100$.

The seventh function is the shifted and rotated Griewank function
with the minimum $f_{7,\min}=-180$ in the domain of $0 \le x_i \le 600$.

The eighth function is Function 23 of the CEC2005 benchmarks \cite{CEC2005},
which is a non-continuous rotated hybrid composition function with the
minimum $f_{8,\min}=360$ in $-5 \le x_i \le 5$.

The ninth function is Function 24 of the CEC2005 benchmarks \cite{CEC2005},
which is a rotated hybrid composition function with $f_{9, \min}=260$ in $-5 \le x_i \le 5$.

The tenth function is Function 25 of the CEC2005 benchmark suite \cite{CEC2005},
which is a rotated hybrid composition function without bounds with the minimum of $f_{10,\min}=260$ in $[2,5]^D$.

\qq{
The next five functions are taken from the CEC2015 benchmark suite \cite{CEC2015a,CEC2015b}. The 11th function is the rotated bent cigar function with the minimum $f_{11, \min}=100$, while the 12th function is the
rotated Discus function with the minimum of $f_{12,\min}=200$.
The 13th function is the shifted and rotated Weierstrass function with $f_{13,\min}=300$,
and the 14th function is the shifted and rotated Schwefel's function with $f_{14,\min}=400$. Finally, the 15th function is the shifted and rotated Katsuura function with $f_{15, \min}=500$. All these functions have variables in the domain of
[-100,100]$^D$ where $D$ is the dimensionality of the functions. }

\subsection{Parameter Settings} \label{Sec-ps-100}

For the implementations, we have used
$n_1=n_2=20$, $n=40$, $r=1$, $m=2$, $N_h=80$, $q=4$ and $w=20$
for the two sets of populations.
For the parameters in the
algorithmic equations, we have used $\a=\b=0.01$, $\lam=1.5$, $p_a=0.25$
and a fixed number of iterations $t_{\max}=1000$ as the stopping criterion.

\qq{In addition,  $D=10$ is used for all the test functions in the first experiment, then $D=50$ is used for the second numerical experiment, while all other parameter values remain the same. For both the standard
cuckoo search (CS) and the proposed MSCS,
we have used $n_{cs}=80$ so that the total numbers of
function evaluations remain the same for all algorithms.
Thus, the fairness of the comparison in terms of function evaluations is ensured.
The parameter setting has been based on preliminary parametric studies in our
own simulation as well as the suggestions in the literature \cite{YangDeb,YangDebRev}.
}

\subsection{Results}

Each algorithm has been run for $100$ trials so as to calculate the best (minimum)
and means of the obtained solutions.
The error is defined as the absolute value of the difference between the best
$f(\x_*)$ found by the algorithm and the true minimum $f_{\min}$(true).
That is \be E_f=|f(\x_*)-f_{\min}(\textrm{true})|. \ee
The numerical results are summarized in
Table~\ref{table-1} where the best corresponds to the minimum of $E_f$
and the mean corresponds to the average value of $E_f$.

From Table 1, we can see that MSCS can obtain better results in all the benchmarks.
The diversity among the cuckoo-host populations in MSCS is higher than those in CS, and the MSCS can be potentially more robust.
This will in general promote the exploration ability of the search process.

\begin{figure}
\centerline{\includegraphics[height=2.5in,width=3.5in]{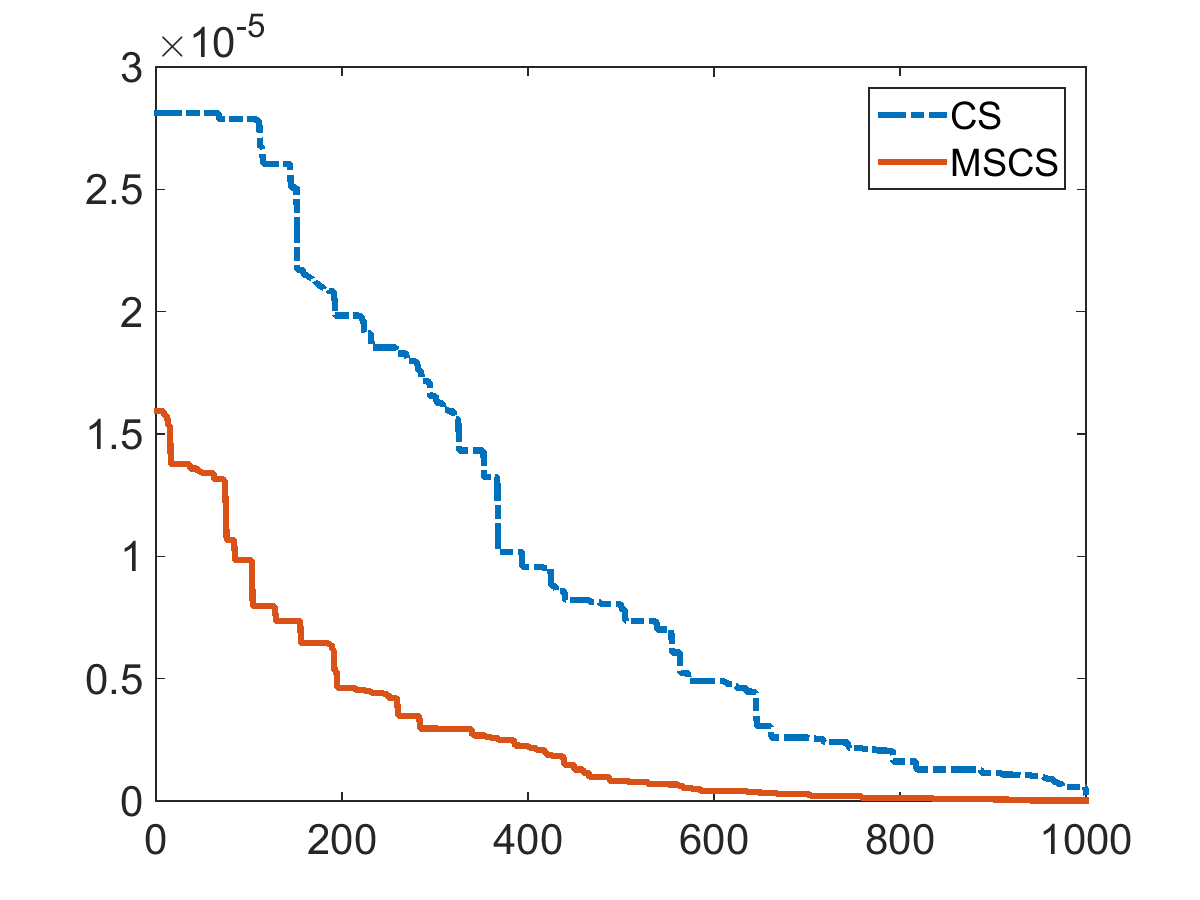}}
\caption{Convergence plot for $f_5$ during iterations.  \label{fig-plot} }
\end{figure}

Another way of looking at the simulation results is to analyze and compare the convergence
behaviour. In fact, MSCS converges faster than CS for all the test functions by tracing
both the minimum objective values found during iterations. For example, for function $f_5$,
its convergence plot is shown in Fig.~\ref{fig-plot} where we can see that MSCS converges
faster even from the very early iterations.

\begin{table}
\begin{center}
\caption{Errors $|f(\x_*)-f_{\min}(\textrm{true})|$ for $D=10$.  \label{table-1} }
\qq{
\begin{tabular}{|c|cc|cc|}
\hline
  & CS & & MSCS & \\ \hline
Function & Best & Mean & Best & Mean \\ \hline
$f_1$ & 2.97E-09 & 1.71E-06 & 2.21E-11 & 3.25E-08  \\
$f_2$ & 2.12E-09 & 1.69E-08 & 1.41E-11 & 5.79E-09 \\
$f_3$ & 7.02E-07 & 5.86E-06 & 3.68E-10 & 2.41E-09 \\
$f_4$ & 3.56E-07 & 2.23E-04 & 8.17E-08 & 7.91E-05 \\
$f_5$ & 4.11E-07 & 5.39E-06 & 1.01E-09 & 5.11E-08 \\
$f_6$ & 1.25E-09 & 2.77E-08 & 7.23E-10 & 5.98E-09 \\
$f_7$ & 2.17E-08 & 5.25E-08 & 2.49E-09 & 5.14E-09 \\
$f_8$ & 4.51E+01 & 7.26E+02 & 1.37E+01 & 9.40E+01 \\
$f_9$ & 2.65E+02 & 7.01E+02 & 3.86E+01 & 2.08E+01 \\
$f_{10}$ & 7.92E+02 & 7.89E+02 & 9.32E+01 & 2.65E+01  \\
$f_{11}$ & 8.14E+02 & 9.01E+02 & 1.27E+02 & 6.32E+02 \\
$f_{12}$ & 2.59E+02 & 6.87E+02 & 3.92E+01 & 7.41E+01 \\
$f_{13}$ & 2.76E+03 & 8.23E+03 & 4.62E+02 & 6.98E+03 \\
$f_{14}$ & 5.89E+03 & 7.54E+03 & 2.51E+03 & 3.42E+03 \\
$f_{15}$ & 2.83E+03 & 2.57E+03 & 1.49E+03 & 4.07E+03 \\
\hline
\end{tabular}  }
\end{center}
\end{table}

\qq{In order to test the proposed algorithm for solving higher-dimensional problems, we have also tested the function benchmarks for $D=50$. In most literature, researchers tend to use higher numbers of iterations for higher values of $D$, typically $t=1000D$, but we have
used the same settings of the parameters as before; that is, $t_{\max}=1000$.
The results are summarized in Table \ref{table-2}.  }

\begin{table}
\begin{center}
\qq{
\caption{Errors $|f(\x_*)-f_{\min}(\textrm{true})|$ for $D=50$.  \label{table-2} }
}
\qq{
\begin{tabular}{|c|cc|cc|}
\hline
  & CS & & MSCS & \\ \hline
Function & Best & Mean & Best & Mean \\ \hline
$f_1$ & 2.21E-06 & 7.43E-07 & 5.04E-09 & 6.37E-07  \\
$f_2$ & 2.83E-08 & 9.74E-07 & 1.42E-09 & 1.75E-07 \\
$f_3$ & 3.35E-06 & 6.21E-05 & 1.91E-07 & 3.22E-08 \\
$f_4$ & 1.94E-05 & 7.73E-01 & 3.89E-04 & 8.79E-01 \\
$f_5$ & 4.51E-05 & 6.12E-03 & 4.81E-06 & 7.33E-06 \\
$f_6$ & 3.31E-07 & 9.17E-06 & 2.92E-07 & 9.05E-06 \\
$f_7$ & 5.27E-06 & 7.51E-06 & 3.88E-07 & 4.71E-06 \\
$f_8$ & 2.83E+03 & 8.12E+03 & 0.91E+03 & 1.38E+03 \\
$f_9$ & 8.62E+03 & 8.90E+03 & 2.01E+03 & 8.92E+03 \\
$f_{10}$ & 5.22E+04 & 9.27E+04 & 1.59E+03 & 8.33E+03  \\
$f_{11}$ & 6.73E+03 & 7.93E+03 & 2.25E+02 & 7.16E+03 \\
$f_{12}$ & 3.12E+03 & 8.89E+03 & 1.37E+03 & 4.70E+03 \\
$f_{13}$ & 2.24E+04 & 7.67E+05 & 2.31E+04 & 5.91E+04 \\
$f_{14}$ & 7.36E+04 & 4.93E+05 & 4.98E+04 & 9.87E+04 \\
$f_{15}$ & 5.05E+04 & 9.28E+04 & 2.05E+03 & 6.69E+04 \\
\hline
\end{tabular}  }
\end{center}
\end{table}

\qq{As we can see from Table 2 that the MSCS obtained better results for almost all functions, except for the shifted and rotated Weierstrass function $f_{13}$
where the two algorithms obtained the same orders of results, but the variation of MSCS is small.
This means that the MSCS not only can produce optimal solutions, but also is sufficiently robust.
 }

\section{Practical Applications}

To test the proposed MSCS algorithm further, we now use five test problems in real-world applications
with diverse properties and nonlinearity. \qq{Three case studies are about designs in engineering and they are mostly mixed integer programming problems. The fourth case study is the parameter estimation problem or an inverse problem, which requires to solve a second-order differential equation to calculate the values of objective function.
The final problem is the data clustering using the well-know Fisher's iris flower data set. }

\qq{It is worth pointing out that these case studies are seemingly simple, but they are quite hard to solve due to high nonlinearity, multimodality and irregular search domains. The pressure vessel problem is also a mixed integer programming problem, which is much harder to solve, compared its continuous counterpart.
}

\subsection{Design of a Spring}
Let us start with a simple but nonlinear problem about the design of spring
under tension or compression \cite{Arora,Coello} from a metal wire. There are three design variables: the wire diameter ($r$), the mean coil diameter ($d$), and the number ($N$) of turns/coils. The objective is to minimize the overall weight of the spring
\be \textrm{minimize } \; f(\x)=(2+N) r^2 d, \ee
subject to nonlinear constraints:
\be g_1(\x)=1- \frac{N d^3}{71785 r^4} \le 0, \quad g_2(\x)=\frac{ d (4 d-r)}{12566 r^3 (d-r)} +\frac{1}{5108 r^2}-1 \le 0, \ee
\be g_3(\x)=1-\frac{ 140.45 r }{d^2 N} \le 0, \quad
g_4(\x)=(d+r) -1.5 \le 0. \ee
Some simple bounds or limits for the design variables are:
\be 0.05 \le r \le 2.0, \quad 0.25 \le d \le 1.3, \quad 2.0 \le N \le 15.0. \ee

Using the proposed MSCS with the same parameter settings given in Section \ref{Sec-ps-100}, \qq{the results of 20 different runs are summarized in Table 3 where comparison is also made.
As we can see, MSCS can obtain the best or the same results as the best results in the literature. }

\begin{table}
\begin{center}
\caption{Comparison of optimal solutions for spring design.}
\begin{tabular}{|c|l|c|}
\hline
Author & Optimal solution & Best objective \\
\hline
Arora \cite{Arora} & (0.053396, 0.399180, 9.185400) & 0.01273 \\ \hline
Coello \cite{Coello} & (0.051480, 0.351661, 11.632201) & 0.01271 \\ \hline
Yang and Deb \cite{YangDeb} & (0.051690, 0.356750, 11.28716) & 0.012665 \\ \hline
Present & (0.051690, 0.356750, 11.28716) & 0.012665 \\
\hline
\end{tabular}
\end{center}
\end{table}

\subsection{Pressure Vessel Design}
A well-known design benchmark is the pressure vessel design problem that has been used by many researchers. This problem concerns the minimization of the overall cost of a cylindrical vessel subject to stress and volume requirements. There are four design variables, including the thickness $d_1$ and $d_2$ for the head and body, respectively, of the vessel, \qq{the inner radius $r$ and the length $W$ of the cylindrical section \cite{Cag,Coello}.  The main objective is
\be \min f(\x)=06224 r W d_1 +1.7781 r^2 d_2 + 19.64 r d_1^2 + 3.1661 W d_1^2, \ee
subject to four constraints:
\be g_1(\x)=-d_1 + 0.0193 r \le 0, \quad g_2(\x)=-d_2 + 0.00954 r \le 0, \ee
\be g_3(\x)=- \frac{4 \pi r^3}{3} - \pi r^2 W -1296000 \le 0, \quad
g_4(\x) =W -240 \le 0. \ee
The inner radius and length are limited to $10.0 \le r, W \le 200.0$. However, the thickness $d_1$ and $d_2$ can only be the integer multiple of a basic thickness of 0.0625 inches. Thus, the simple bounds for thickness are
\be 1 \times 0.0625 \le d_1, d_2 \le 99 \times 0.0625. \ee
This is a mixed integer programming because two variables are discrete and the other two variables are continuous. }

\begin{table}
\begin{center}
\caption{Comparison of optimal solutions for pressure vessel design.}
\begin{tabular}{|c|l|c|}
\hline
Author & Optimal solution & Best objective \\
\hline
Cagnina et al. \cite{Cag} & (0.8125, 0.4375, 42.0984, 176.6366) & 6059.714 \\ \hline
Coello \cite{Coello} & (0.8125, 0.4375, 42.3239, 200.0) & 6288.7445 \\ \hline
Yang et al. \cite{YangV} & (0.8125, 0.4375, 42.0984456, 176.6365959) & 6059.714 \\ \hline
Present & (0.8125, 0.4375, 42.0984456, 176.6366) & 6059.714 \\
\hline
\end{tabular}
\end{center}
\end{table}

\qq{Using the same parameter settings as before, the results of 20 independent runs are summarized and compared in Table 4. In fact, all these algorithms can find the global optimal solution as found by Yang et al. \cite{YangV}. }

\subsection{Speed Reducer Design}
The speed reducer design is an engineering design benchmark, which has 7 design variables such as the face width of the gear, number of teeth, and diameter of the shaft and others \cite{Golinski}. All these variables can take continuous values, except for $x_3$ which is an integer.

The objective to minimize the cost function
\[ f(\x)=0.7854 \Big[x_1 x_2^2 (3.3333 x_3^2 +14.9334x_3-43.0934 ) + (x_4 x_6^2+x_5 x_7^2) \Big] \]
\be -1.508 x_1 (x_6^2+x_7^2)+7.4777 (x_6^3+x_7^3), \ee
subject to 11 constraints:
\be g_1(\x)=\frac{27}{x_1 x_2^2 x_3}-1 \le 0, \quad
g_2(\x)=\frac{397.5}{x_1 x_2^2 x_3^2} -1 \le 0, \ee
\be g_3(\x)=\frac{1.93 x_4^3}{x_2 x_3 x_6^4}-1 \le 0, \quad
g_4(\x)=\frac{1.93 x_5^3}{x_2 x_3 x_7^4}-1 \le 0, \ee
\be g_5(\x) =\frac{1.0}{110 x_6^3} \sqrt{(\frac{745.0 x_4}{x_2 x_3})^2+16.9 \times 10^6}-1 \le 0, \ee
\be g_6(\x) =\frac{1.0}{85 x_7^3} \sqrt{(\frac{745.0 x_5}{x_2 x_3})^2 +157.5 \times 10^6 }-1 \le 0, \ee
\be g_7(\x)=x_2 x_3-40 \le 0,
\quad g_8(\x)=5 x_2 -x_1 \le 0, \ee
\be g_9(\x) =x_1 -12 x_2 \le 0, \quad g_{10}(\x)=(1.5 x_6 +1.9) - x_4 \le 0, \ee
\be g_{11}(\x) =(1.1 x_7 + 1.9) - x_5 \le 0. \ee
In addition, the simple bounds for the variables are:
$2.6 \le x_1 \le 3.6$, $0.7 \le x_2 \le 0.8$, $17 \le x_3 \le 28$ (integers only), $7.3 \le x_4 \le 8.3$, $7.8 \le x_5 \le 8.4$, $2.9 \le x_6 \le 3.9$,
and $5.0 \le x_7 \le 5.5$.

\begin{table}
\begin{center}
\caption{Comparison of optimal solutions for the speed reducer problem.}
\begin{tabular}{|c|l|l|}
\hline
Author & Optimal solution & Best objective \\
\hline
Akhtar et al. \cite{Akhtar} & (3.5061, 0.7, 17, 7.549, 7.8593, 3.3656,5.28977) & 3008.08 \\
\hline
Cagnita et al. \cite{Cag} & (3.5, 0.7, 17, 7.3, 7.8, 3.350214, 5.286683) & 2996.348165 \\
\hline
Yang \& Gandomi \cite{YangBATEO} & (3.5, 0.7, 17, 7.3, 7.71532, 3.35021, 5.2875 & 2994.467 \\
\hline
Present & (3.5, 0.7, 17, 7.3, 7.8, 3.34336449, 5.285351) & 2993.749589 \\
\hline

\end{tabular}
\end{center}
\end{table}

\qq{The results of the 20 independent runs are summarized and comparison has been made in Table 5. As we can see, MSCS obtained the best result. Since there is no literature about the analysis of this problem and we do not know what the global best solution should be, we can only say that $2993.749589$ is the best result achieved so far. }

\subsection{Parameter Identification of Vibrations}
For a simple vibration problem,
there are two unknown parameters $\mu$ and $\nu$ to be estimated
from the measurements of the vibration amplitudes. The governing equation is
\be \frac{d^2 y(t)}{dt^2} + \mu \frac{dy(t)}{dt}+\nu y(t)=40 \cos (3 t). \label{ODE-100} \ee
In general, the solution is a damped harmonic motion. However, for fixed $\mu=4$
and $\nu=5$ with an initial values of $y(0)=0$ and $y'(0)=0$, there is an analytical solution \cite{YangMaths}, which can be written as
\be y(t)=e^{-2t} [\cos(t)-7 \sin (t)]+3 \sin(3 t)-\cos (3 t). \ee

For a real system with a forcing term $40 \cos (3 t)$, we do not know the parameters, but
the vibrations can be measured.
For example, in an experiment, there are $N=11$ measurements as shown in Table \ref{table-6}.
\begin{table}
\begin{center}
\caption{Measured response of a simple vibration system. \label{table-6}}
\begin{tabular}{|l|lllllllllll|}
\hline
$t$ & 0.00 & 0.20 & 0.40 & 0.60 & 0.80 & 1.00 & 1.20 & 1.40 & 1.60 & 1.80 & 2.00  \\ \hline
$y_d(t)$ & 0.00 & 0.59 & 1.62 &  2.21 & 1.89 & 0.69 & -0.99 & -2.53 & -3.36 &
-3.15 & -1.92 \\ \hline
\end{tabular}
\end{center}
\end{table}

The task is to estimate the values of the two parameters. However, one of the main challenges is that the calculation of the objective function that is defined as the sum of errors squared. That is
\be f(\x)=\sum_{i=1}^{N} (y_{i,{\rm predicted}} - y_{i,d})^2, \ee
where the predicted $y(t)$ has to be obtained by solving the second-order ordinary differential equation (\ref{ODE-100}) numerically and iteratively for every guessed set of $\mu$ and $\nu$.

Using the MSCS with a population of $40$ cuckoos and the same parameter setting given in Section \ref{Sec-ps-100}. We have run the algorithm for 20 times, and the mean values of estimate parameters are: $\mu_*=4.025$ and $\nu_*=4.981$, which are very close to the true values of $\mu=4.000$ and $\nu=5.000$.

\subsection{Iris Classification}
To test the MSCS algorithm further, we use it to solve the classification problem of the well-known Fisher's Iris flower data set. This data set has 150 data points or instances with 4 attributes and 3 distinct classes \cite{Duda}. We use the data from the UCI Machine Learning Repository\footnote{http://archive.ics.uci.edu/ml/datasets/Iris}.

We have encoded the centres of the clusters as the solution vectors  so as to minimize the overall intra-clustering distances. Although the parameter settings are the same as before, the number of iterations is limited to 100 so as to be comparable with the results from the literature \cite{Binu}. The best values obtained are compared with the results obtained by other methods such as the hybrid k-means and PSO approach \cite{Kao}, multiple kernel based fuzzy c-means with cuckoo search \cite{Binu}, and k-means with improved feature based cluster centre initialization algorithm (CCIA) \cite{Khan}.

\qq{The optimization results are summarized in Table \ref{table-5}. As we can see, MSCS obtained the best result which signifies an improvement over the best results obtained by multiple kernel fuzzy c-means based cuckoo search approach (MKF-cuckoo) \cite{Binu}. }

\begin{table}
\begin{center}
\caption{Accuracy comparison for Iris data set. \label{table-5}}
\begin{tabular}{|c|c|c|}
\hline
Method & Author & accuracy \\ \hline
K-means and PSO & Kao et al. \cite{Kao} & $89.3\%$ \\ \hline
MKF-Cuckoo & Binu et al. \cite{Binu} & $95.0\%$ \\ \hline
K-means & Khan \& Ahmad \cite{Khan} & $76.4\%$ \\ \hline
K-means with CCIA & Khan \& Ahmad \cite{Khan} & $88.7\%$  \\ \hline
MSCS & this paper & $97.1\%$ \\ \hline
\end{tabular}
\end{center}
\end{table}

The results and simulation we have obtained so far are indeed encouraging. Obviously, we will carry out more thorough evaluations of the proposed approach in the future work. So let us summarize the work we have done so far in this paper.

\section{Discussions}

The original cuckoo search has been extended to capture more realistic characteristics of
cuckoo-host co-evolution systems. We have thus developed a multi-species cuckoo search for solving optimization problems using multiple cuckoo species competing and co-evolving with host species. We then tested the proposed approach using a set of 15 function benchmarks to show that the proposed algorithm can indeed work well. Preliminary results suggest that MSCS can have a higher convergence rate and obtain better results in general. In addition, we have used 5 different case studies from engineering designs, parameter estimation and data clustering to further test the proposed algorithm. Our simulation results and subsequent comparison have shown that the MSCS can indeed find the optimal solutions that are either better and comparable with the results obtained by other methods.

\qq{The essence of multi-species co-evolution is to use more than one species so as to see how different species interact and compete. In the present studies, we have just used $m=2$ species for the cuckoo-host co-evolution. Future work will focus on more than two species and more detailed parametric studies using different numbers of species $m$ with varied population sizes. In addition, it would gain more insight by tuning the key parameters in the algorithm to see how they may affect the overall performance of the algorithm. }

\qq{Furthermore, in the real-world cuckoo-host co-evolution systems, there are multiple cuckoo species interacting with multiple host species, which can have much more complex behaviour and characteristics. The current approach with the preliminary tests consists of only a single host species with multiple cuckoo species. A possible extension can be the multiple host bird species compete and co-evolve with multiple cuckoo species. For example, the common cuckoos can lay eggs in many different host species including garden warblers and reed warblers \cite{Payne}.
The number of eggs laid by cuckoos and inside nests can be random.
It can also be useful to carry out further tests of this algorithm using a more extensive set of benchmarks and real-world case studies. }

\qq{
{\bf Acknowledgement}: The authors would like to thank the anonymous reviewers for their constructive comments.   }



\begin{thebibliography}{99.}

\bibitem{Akhtar} Akhtar S, Tai K, Tay T. A socio-behavioural simulation model for engineering design optimization. Engineering Optimization 2002; 34(4): 341--454.

\bibitem{Arora} Arora JS. Introduction to Optimum Design. New York: McGraw-Hill; 1989.

\bibitem{Bhar}
Bhargava V, Fateen, SEK, Bonilla-Petriciolet A. (2013).
Cuckoo search: a new nature-inspired optimization method
for phase equilibrium calculations. Fluid Phase Equilibria 2013;  337:191--200.

\bibitem{Binu} Binu D, Selvi M, Aloysius G. MKF-cuckoo: hyrbidization of cuckoo search and multiple kernel-based fuzzy c-means algorithm. AASRI Procedia 2013; 4: 243--249.

\qq{\bibitem{Blackwell} Blackwell T and Branke J, Multi-swarm optimization in dynamic environments, in: Applications of Evolutionary Computing, EvoWorkshops 2004, Lecture Notes in Computer Science, Vol. 3005, Springer, Berlin, 2004; 489--500.
}

\bibitem{Cag} Cagnina LC, Esquivel SC, Coello Coello CA. Solving engineering optimization problems with the simple constrained particle swarm optimizer. Informatica 2008; 32:319--326.


\bibitem{Chand}
Chandrasekaran K,  Simon SP.  Multi-objective scheduling problem: hybrid appraoch using fuzzy assisted cuckoo search algorithm. Swarm and Evolutionary Computation 2012; 5(1): 1--16.

\bibitem{CEC2015a}
\qq{
Chen Q, Liu B, Zhang Q, Suganthan PN, Qu BY, Problem definition and evaluation criteria for CEC2015 special session and competition on bound constrained single-objective computationally expensive numerical optimization, Technical Report, Commputational Intelligence Laboratory, Zhengzhou University, China and Technical Report, Nanyang Technology Univesity, Singapore, Nov. 2014.
}

\bibitem{Coello} Coello Coello CA. Use of a self-adaptive penalty approach for engineering optimization problems. Computers in Industry 2000; 41: 113-127.


\bibitem{Davi} Davies NB and Brooke ML. Co-evolution of the cuckoo and its hosts. Scientific American 1991; 264(1):92-98.


\bibitem{Davi2} Davies NB. Cuckoo adaptations: trickery and tuning.
Journal of Zoology 2011;  284(1): 1--14.


\bibitem{Dhiv} Dhivya M and Sundarambal M. Cuckoo search for data gathering in wireless sensor networks. Int. J. Mobile Communications 2011; 9(4): 642-656.

\qq{\bibitem{Dubey} Dubey HM, Pandit M, Panigrahi BK, A biologically inspired modified flower pollination algorithm for solving dispatch problems in modern power systems. Cognitive Computation 2015; 7(5): 594--608. }

\bibitem{Duda} Duda RO, Hart PE. Pattern Classification and Scene Analysis.
New York: John Wiley and Sons; 1973.

\bibitem{Durgun}
Durgun I, Yildiz AR. Structural design optimization of vehicle components using cuckoo search algorithm.  Materials Testing 2012; 3(3):  185-188.


\bibitem{Fister} Fister Jr I, Fister D, Fister I. A comprehensie review of cuckoo search: variants and hybrids. Int. J. Mathematical and Numerical Optimisation 2013; 4(4): 387--409.


\bibitem{Gandomi}
Gandomi AH, Yang XS, Alavi AH. Cuckoo search algorithm: a metaheuristic approach to solve structural optimization problems.  Engineering with Computers 2013; 29(1): 17--35.


\bibitem{Golinski} Golinski J. An adaptive optimization system applied to machine synthesis. Mech. Mach. Theory 1973; 8(4): 419--436.

\bibitem{Kao} Kao Y-T, Zahara E, Kao I-W. A hybridized approach to data clustering. Expert Systems with Applications 2008; 34(3): 1754--1762.


\bibitem{Khan} Khan SS, Ahmad A. Cluster center initialization algorithm for k-means clustering. Pattern Recognition Letters 2004; 25(11): 1393--1302.

\bibitem{Kruger} Kr\"uger O, Sorenson MD, Davies NB.
Does co-evolution promote species richness in parasitic cuckoos?  Proc. Roy. Soc. B 2009; 276(1674): 3871--3879.

\bibitem{Mishra} Mishra SK. Global optimization of some difficult benchmark functions by host-parasite co-evolutionary algorithm. Economics Bulletin 2013; 33(1): 1--18.


\bibitem{Mlak} Mlakar U, Fister Jr I, Fister I. Hybrid self-adaptie cuckoo search for global optimization. Swarm and Evolutionary Computation 2016; 29: 47--72.

\bibitem{Moham} Mohamad AB, Zain AM, Bazin NEN, Cuckoo search algorithm for optimization problems -- A literature review and its applications.
    Applied Artificial Intelligence 2014; 28 (5): 419--448.



\bibitem{Mora}
Moravej Z, Akhlaghi A. A novel approach based on cuckoo search for DG allocation in distribution network. Electrical Power and Energy Systems 2013; 44(1): 672--679.

\bibitem{Pare} Pare S, Kumar A, Bajaj V, Singh GK. A multilevel color image segmentation technique based on cuckoo search algorithm and energy curve, Applied Soft Computing 2016; 47:76--102.

\bibitem{Payne} Payne RB. The Cuckoos. Oxford: Oxford University Press; 2005.

\bibitem{Pav} Pavlyukevich I. L\'evy flights, non-local search and simulated annealing. J. Computational Physics 2007;  226(2):1830-1844.


\bibitem{Pereira} Pereira LAM, Rodrigues D, Almeida TNS, Ramos CCO, Souza AN, Yang XS, Papa JP. A binary cuckoo search and its application for feature selection. in: Cuckoo Search and Firefly Algorithm. Studies in Computational Intelligence 2013; Vol.516, pp.141--154.

\bibitem{CEC2015b}
\qq{
Qu BY, Liang JJ, Wang ZY, Chen Q, Suganthan PN, Novel benchmark functions for continuous multimodal optimization with comparative results, {\it Swarm and Evolutionary Computation} 2016; 26(1): 23-34.
}


\bibitem{Santos} Santos CAG, Freire PKMM, Mishra SK. Cuckoo search via L\'evy fligths for optimization of a physically-based runoff-erosion model. Journal of Urban and Environmental Engineering 2012; 6(2): 123--131.

\bibitem{Shehab} Shehab M, Khader AT, Al-Betar MA. A survy on applications and variants of the cuckoo search algorithm. Applied Soft Computing 2017;
    https://doi.org/10.1016/j.asoc.2017.02.034


\qq{\bibitem{Sidd} Siddique N, Adeli H, Nature-inspired chemical reaction optimisation algorithms. Cognitive Computation 2017; 9: 411-422. }

\bibitem{CEC2005} Suganthan PN, Hansen N, Liang JJ,
Deb K, Chen YP, Auger A, Tiwari S. Problem definitions and evaluation criteria for the CEC2005 special session on real-parameter optimization, Technical Report of Nanyang Technological University, Singapore and KanGAL Report, IIT Kanpur, India; 2005.

\bibitem{Valian} Valian E, Mohanna S, Tavakoli S.
Improved cuckoo search algorithm for feedforward neural network training.
Int. J. Articial Intelligence and Applications 2011;  2(3): 36--43.


\bibitem{Walton} Walton S, Hassan O, Morgan K, Brown MR.
Modified cuckoo search: a new gradient free optimization algorithm.
Chaos, Solitons \& Fractals 2011; 44(9): 710-718.

\bibitem{Wang} Wang GG, Deb S, Gandomi AH, Zhang ZJ, Alavi AH.
Chaotic cuckoo search. Soft Computing 2016; 20(9): 3349--62.

\bibitem{Wong} Wong PK, Wong KI, Vong CM, Cheung CS.
Modeling and optimization of biodiesel energy performance using
kernel-based extreme learning machine and cuckoo search.
Renewable Energy 2015; 74: 640--647.

\bibitem{Woz} Wo\'zniak M, Polap D, Napoli C, Tramontana E. Graphic object feature extraction system based on cuckoo search algorithm. Expert Systems with Applications 2016; 66: 20--31.

\qq{\bibitem{Wu} Wu TQ, Yao M, Yang, JH. Dophin swarm extreme learning machine.
Cognitive Computation 2017; 9(2): 275--284. }


\bibitem{YangDeb} Yang XS, Deb S. Cuckoo search via L\'evy flights,
 in: Proc. of World Congress on Nature \& Biologically Inspired Computing (NaBic 2009), India, IEEE Publications, USA. 2009; pp. 210--214

\bibitem{YangDeb2010} Yang XS, Deb S. Engineering optimization by cuckoo search. Int. J. Math. Modelling Num. Optimisation 2010; 1(4): 330--343.

\bibitem{YangDebRev} Yang XS, Deb S. Cuckoo search: recent advances and applications. Neural Computing and Applications 2014; 24(1): 169-174.

\bibitem{YangBATEO} Yang XS, Gandomi AH. Bat algorithm: a novel approach for
global engineering optimization. Engineering Computations 2012; 29(5): 464--483.

\bibitem{YangCSFA} Yang XS. Cuckoo Search and Firefly Algorithm: Theory and Applications. Studies in Computational Intelligence, Vol. 516, Springer; 2014.

\bibitem{YangDeb2013} Yang XS, Deb S.  Multiobjective cuckoo search for design optimization. Computers and Operations Research 2013; 40(6): 1616-1624.

\bibitem{YangV} Yang XS, Huyck C, Karamanoglu M, Khan N. True global optimality of the pressure vessel design problem: a benchmark for bio-inspired optimisation algorithms. Int. J. Bio-Inspired Computation 2013; 5(6): 329--335.

\bibitem{YangMaths} Yang XS. Engineering Mathematics with Examples and Applications. London: Academic Press; 2017.

\bibitem{Yao} Yao X, Liu Y, Lin G. Evolutionary programming made faster. IEEE Trans. Evol. Computation 1999; 3(2): 82-102.

\bibitem{Yildiz} Yildiz AR.
Cuckoo search algorithm for the selection of optimal machine parameters in milling operations. Int. J. Adv. Manuf. Technol. 2013;  64(1): 55--61.

\bibitem{Zamani} Zamani AA, Tavakoli S, Etedali S. Fractional order PID control design for semi-active control of smart base-isolated structures: a multi-objective cuckoo search approach. ISA Tractions 2017; 67: 222-232.

\bibitem{Zheng}
Zheng HQ, Zhou Y. A novel cuckoo search optimization
algorithm based on Gauss distribution. J. Computational Information Systems 2012; 8(10): 4193--4200.

\bibitem{Zine} Zineddube M. Vulnerabilities and mitigation techniques toning in the cloud: a cost and vulnerablities coverage optimization approach using cuckoo search algorithm with L\'evy flights. Computers \& Security 2015;
    48: 1--18.



\end{thebibliography}
\end{document}